\documentclass[11pt,a4paper]{article}
\usepackage[hyperref]{emnlp2020}
\pdfoutput=1
\usepackage{times}
\usepackage{latexsym}

\usepackage{times}
\usepackage{latexsym}
\usepackage{amsmath}
\usepackage{url}
\usepackage{amssymb}
\usepackage{amsfonts}
\usepackage{graphicx}
\usepackage{tabularx}
\usepackage{multirow}
\usepackage{arydshln}
\usepackage{mathtools,nccmath}

\usepackage{enumitem}
\aclfinalcopy % Uncomment this line for the final submission

\title{BERTweet: A pre-trained language model for English Tweets}

\author{Dat Quoc Nguyen$^1$, Thanh Vu$^{2,}$\thanks{\ \ Most of the work done
when Thanh Vu was at the Australian e-Health Research Centre, CSIRO, Australia. }   \and Anh Tuan Nguyen$^{3,}$\thanks{\ \ Work done during internship at  VinAI Research.} \\
  $^1$VinAI Research, Vietnam; $^2$Oracle Digital Assistant, Oracle, Australia; $^3$NVIDIA, USA\\
   \tt{\normalsize v.datnq9@vinai.io; thanh.v.vu@oracle.com; tuananhn@nvidia.com}}

\date{}

\begin{document}
\maketitle
\begin{abstract}
We present \textbf{BERTweet}, the \emph{first} public large-scale pre-trained language model  for English Tweets. Our BERTweet, having the same architecture as BERT\textsubscript{base} \citep{devlin-etal-2019-bert}, is trained using the RoBERTa pre-training procedure \citep{RoBERTa}. 
Experiments show that BERTweet outperforms strong baselines RoBERTa\textsubscript{base} and XLM-R\textsubscript{base} \citep{conneau2019unsupervised}, producing better performance results than the previous state-of-the-art models on three Tweet NLP tasks: Part-of-speech tagging, Named-entity recognition and text classification. We release BERTweet under the MIT License to facilitate future research and  applications on Tweet data. Our BERTweet is available at: %\url{anonymous-url}.
\url{https://github.com/VinAIResearch/BERTweet}.
\end{abstract}

\section{Introduction}\label{sec:intro}

The language model BERT \citep{devlin-etal-2019-bert}---the Bidirectional Encoder Representations from Transformers \citep{NIPS2017_7181}---and its variants 
have successfully helped produce new state-of-the-art performance results for various NLP tasks. Their success has largely covered the common English domains such as Wikipedia, news and books. For specific domains such as biomedical or scientific, we could retrain a domain-specific model using the BERTology architecture \cite{beltagy-etal-2019-scibert,btz682,dontstoppretraining2020}.

Twitter has been one of the most popular micro-blogging platforms where users can share real-time information related to all kinds of topics and events. The enormous and plentiful Tweet data has been proven to be a widely-used and real-time source of information in various important analytic tasks \cite{GhaniHHA19}. 
Note that the characteristics of Tweets are generally different from those of traditional written text such as Wikipedia and news articles, due to the typical short length of Tweets and frequent use of informal grammar as well as irregular vocabulary e.g. abbreviations, typographical errors and hashtags \citep{eisenstein-2013-bad,2414425.2414430}. Thus this might lead to a challenge in applying existing language models pre-trained on large-scale conventional text corpora with formal grammar and regular vocabulary to handle text analytic tasks on Tweet data. To the best of our knowledge, there is not an existing language model pre-trained on a large-scale corpus of English Tweets. 

To fill the gap, we train the \textit{first} large-scale language model for English Tweets using a 80GB corpus of 850M English Tweets. Our model uses the BERT\textsubscript{base} model configuration, trained based on the RoBERTa pre-training procedure \citep{RoBERTa}. 
We evaluate our model and compare it with strong competitors, i.e. RoBERTa\textsubscript{base} and XLM-R\textsubscript{base} \citep{conneau2019unsupervised}, on three downstream Tweet NLP tasks:  Part-of-speech (POS) tagging, Named-entity recognition (NER) and text classification. Experiments show that our model outperforms  RoBERTa\textsubscript{base} and XLM-R\textsubscript{base} as well as the previous  state-of-the-art (SOTA) models on 
all these tasks. Our contributions are as follows:

\begin{itemize}[leftmargin=*]
\setlength\itemsep{-1pt}
    \item We present the {first} large-scale  pre-trained language model for English Tweets.
    
    \item Our model does better than its competitors RoBERTa\textsubscript{base} and XLM-R\textsubscript{base} and outperforms previous SOTA models on three downstream Tweet NLP tasks of POS tagging, NER and text classification, thus confirming the effectiveness of the large-scale and domain-specific language model pre-trained for English Tweets.
    
    \item We also provide the first set of experiments investigating whether  a commonly used approach of applying lexical normalization dictionaries  on Tweets \cite{han-etal-2012-automatically} would help improve the performance of the pre-trained language models  on the downstream tasks.
    
    \item We publicly release our model under the name {BERTweet} which can be used with   \texttt{fairseq}  \citep{ott2019fairseq} and \texttt{transformers} \cite{Wolf2019HuggingFacesTS}. We hope that BERTweet can serve as a strong baseline for future  research and applications of Tweet analytic tasks.
\end{itemize}

\section{BERTweet} \label{sec:bertweet}

In this section, we outline the architecture, and describe the pre-training data and optimization setup that we use for BERTweet.

\subsubsection*{Architecture} 
Our  BERTweet uses the same  architecture  as BERT\textsubscript{base}, which is trained with a masked language modeling objective  \citep{devlin-etal-2019-bert}. BERTweet pre-training procedure is based on RoBERTa \citep{RoBERTa} which optimizes the BERT pre-training approach for more robust performance. Given the widespread usage of BERT and RoBERTa, we do not detail the architecture here. See \newcite{devlin-etal-2019-bert} and \newcite{RoBERTa} for more details. 

\subsubsection*{Pre-training data}
We use an 80GB pre-training dataset of uncompressed texts, containing 850M Tweets (16B word tokens). Here, each Tweet consists of at least 10 and at most 64 word tokens. In particular, this dataset is a concatenation of two corpora: 

\begin{itemize}[leftmargin=*]
\setlength\itemsep{-1pt}

\item We first download the general Twitter Stream grabbed by the Archive Team,\footnote{\url{https://archive.org/details/twitterstream}} containing 4TB of Tweet data streamed from 01/2012 to 08/2019 on Twitter. To identify English Tweets, we employ the language identification component of fastText \citep{joulin-etal-2017-bag}.  %\footnote{\url{https://fasttext.cc/docs/en/language-identification.html}} 
We tokenize those English Tweets using ``TweetTokenizer'' from the NLTK toolkit \citep{bird_natural_2009} and use the \texttt{emoji} package to translate emotion icons into text strings (here, each icon is referred to as a word token).\footnote{\url{https://pypi.org/project/emoji}}  We also normalize the Tweets by converting user mentions and web/url links into special tokens \texttt{@USER} and \texttt{HTTPURL}, respectively. We filter out retweeted Tweets and the ones shorter than 10 or longer than 64 word tokens. This pre-process results in the first corpus of 845M English Tweets. 

\item We also stream Tweets related to the COVID-19 pandemic, %\footnote{\url{https://www.who.int/emergencies/diseases/novel-coronavirus-2019}} 
available from 01/2020 to 03/2020.\footnote{We collect Tweets containing at least one of 11 COVID-19 related keywords, e.g. covid19, coronavirus, sars-cov-2.} We apply the same data pre-process step as described above, thus resulting in the second corpus of 5M English Tweets.

\end{itemize}

We then apply \texttt{fastBPE} \citep{sennrich-etal-2016-neural} to segment all 850M Tweets with subword units, using a vocabulary of 64K subword types. On average there are 25 subword tokens per Tweet.

\subsubsection*{Optimization}
We utilize the RoBERTa implementation in the \texttt{fairseq} library \citep{ott2019fairseq}. We set a maximum sequence length at  128, thus generating  850M $\times$ 25 / 128 $\approx$ 166M sequence blocks. % (i.e. 25 subword tokens per sequence on average). 
Following \newcite{RoBERTa}, we optimize the model using Adam \citep{KingmaB14}, and use a batch size of 7K across 8 V100 GPUs (32GB each) and a peak learning rate of 0.0004. We pre-train BERTweet for 40 epochs   in  about {4} weeks (here, we use the first 2 epochs for warming up the learning rate), equivalent to 166M $\times$ 40 / 7K $\approx$ 950K training steps.

%\begin{table}[!t]
%    \centering
%    \resizebox{7.5cm}{!}{
%    \begin{tabular}{l|l|l|l}
%    \hline
%    \textbf{Dataset}  & \textbf{\#training} & \textbf{\#valid} & \textbf{\#test} \\
%    \hline
%    Ritter11-T-POS & 562 & 117 & 118\\
%    ARK-Twitter & 1326 & 492 & 546\\
%    \textsc{Tweebank-v2} & 1635 & 708& 1201\\
%    %\textsc{Tweebank-v2}+cWT & 13657 & 708 & 1201\\
%    \hline
%    WNUT16 & 2393 & 999 & 3844 \\
%    WNUT17  & 3392 & 1008 & 1257\\
%    \hline
%    SemEval-17-Task-4A & 48215 & 5357 & 12285 \\
%    SemEval-18-Task-3A & 3452 & 384 & 785  \\
%    \hline
%    \end{tabular}
%    }
%    \caption{Statistics of the downstream task datasets. Here, ``\#training'', ``\#valid''  and  ``\#test'' denote the number of Tweets in the training, validation and test sets, respectively. }
%    \label{tab:data}
%\end{table}
 
\section{Experimental setup}

We evaluate and compare the performance of BERTweet with strong baselines on three downstream NLP tasks of POS tagging, NER and text classification, using benchmark Tweet datasets.

\subsubsection*{Downstream task datasets} 
For POS tagging, we use three datasets Ritter11-T-POS %\footnote{\url{https://github.com/aritter/twitter_nlp/blob/master/data/annotated/pos.txt}} 
\cite{ritter-etal-2011-named}, ARK-Twitter\footnote{\url{https://code.google.com/archive/p/ark-tweet-nlp/downloads} (twpos-data-v0.3.tgz) } 
\citep{gimpel-etal-2011-part,owoputi-etal-2013-improved} and \textsc{Tweebank-v2}\footnote{\url{https://github.com/Oneplus/Tweebank}} 
\citep{liu-etal-2018-parsing}. For NER, we employ datasets from the WNUT16 NER shared task 
%\footnote{\url{https://noisy-text.github.io/2016/ner-shared-task.html}}
\citep{strauss-etal-2016-results} and the WNUT17 shared task 
%\footnote{\url{https://noisy-text.github.io/2017/emerging-rare-entities.html}} 
on novel and emerging entity recognition \citep{derczynski-etal-2017-results}. For text classification, we employ the 3-class sentiment analysis dataset from the SemEval2017 Task 4A 
%\footnote{\url{http://alt.qcri.org/semeval2017/task4/}} 
\citep{rosenthal-etal-2017-semeval} and the 2-class irony detection dataset from the SemEval2018 Task 3A 
%\footnote{\url{https://competitions.codalab.org/competitions/17468}}
\citep{van-hee-etal-2018-semeval}.

For Ritter11-T-POS, we employ a 70/15/15 training/validation/test pre-split available from \newcite{gui-etal-2017-part}.\footnote{\url{https://github.com/guitaowufeng/TPANN}}
ARK-Twitter contains two files \texttt{daily547.conll} and \texttt{oct27.conll} in which  \texttt{oct27.conll} is further split into files \texttt{oct27.traindev} and \texttt{oct27.test}. Following \newcite{owoputi-etal-2013-improved} and \newcite{gui-etal-2017-part}, we employ \texttt{daily547.conll} as a test set. In addition, we use \texttt{oct27.traindev} and \texttt{oct27.test} as training and validation sets, respectively. For the \textsc{Tweebank-v2}, WNUT16 and WNUT17 datasets, we use their existing training/validation/test split. The SemEval2017-Task4A and SemEval2018-Task3A datasets are provided with training and test sets only (i.e. there is not a standard split for validation), thus we sample 10\% of the training set for validation and use the remaining 90\% for training. % See the Appendix for the statistics of the experimental datasets.
%\footnote{Those two Tweet classification datasets are then tokenized by utilizing ``TweetTokenizer'' from the NLTK toolkit.}  

We use a ``soft'' normalization strategy to all of the experimental datasets by translating word tokens of user mentions and web/url links into special tokens \texttt{@USER} and \texttt{HTTPURL}, respectively, and converting emotion icon tokens into corresponding strings. We also apply a ``hard''  strategy by further applying lexical normalization dictionaries  \cite{TYPO2010,liu-etal-2012-broad,han-etal-2012-automatically} to normalize word tokens in Tweets. %This ``hard'' strategy is to examine whether  the commonly used lexical normalization approach for Tweets  would help improve the performance on the downstream tasks regarding pre-trained language models.

% Table \ref{tab:data} presents the statistics of the benchmark datasets  for downstream task evaluation. 
 
\subsubsection*{Fine-tuning}

Following \newcite{devlin-etal-2019-bert}, for POS tagging and NER, we append a linear prediction layer on top of the last Transformer layer of BERTweet with regards to the first subword of each word token, while for text classification we append a linear prediction layer on top of the pooled output. 

We employ the \texttt{transformers} library \cite{Wolf2019HuggingFacesTS} to independently fine-tune BERTweet for each task and each dataset in 30 training epochs. We use AdamW   \citep{loshchilov2018decoupled} with a fixed learning rate of 1.e-5 and a batch size of 32 \citep{RoBERTa}. We compute the task performance after each training epoch on the validation set  (here, we apply early stopping when no improvement is observed after 5 continuous epochs), and select the best model checkpoint to compute the performance score on the test set. 

We repeat this fine-tuning process 5 times with different random seeds, i.e. 5 runs for each task and each dataset. We report each final test result as an average over the test scores from the 5 runs.

\subsubsection*{Baselines}

Our main competitors are the pre-trained language models RoBERTa\textsubscript{base} \citep{RoBERTa} and XLM-R\textsubscript{base} \citep{conneau2019unsupervised}, which have the same architecture configuration as our BERTweet. In addition, we also evaluate the pre-trained RoBERTa\textsubscript{large} and XLM-R\textsubscript{large} although it is not a fair comparison due to their significantly larger model configurations. 

The pre-trained RoBERTa is a strong language model for English, learned from 160GB of texts covering books, Wikipedia, CommonCrawl news, CommonCrawl stories, and web text contents. XLM-R  is a cross-lingual variant of RoBERTa, trained on a 2.5TB multilingual corpus which contains 301GB of English CommonCrawl texts. %XLM-R is the recent best  multilingual language model, which also outperforms RoBERTa and  obtains state-of-the-art (SOTA) results in multiple English NLP tasks. 

We fine-tune RoBERTa and XLM-R using the same fine-tuning approach  we use for BERTweet.

\begin{table}[!t]
    \centering
\resizebox{7.5cm}{!}{
\setlength{\tabcolsep}{0.25em}
    \begin{tabular}{ll|c|c|c|c|c|c}
    \hline
       \multicolumn{2}{c|}{\multirow{2}{*}{\bf Model}} & \multicolumn{2}{c|}{\textbf{Ritter11}} & \multicolumn{2}{c|}{\textbf{ARK}}& \multicolumn{2}{c}{\textbf{\textsc{TB-v2}}} \\
         \cline{3-8} 
         & & soft & hard & soft & hard & soft & hard\\
         \hline
         \multirow{5}{*}{\rotatebox[origin=c]{90}{Our results}} 
        &  RoBERTa\textsubscript{large} & 91.7    & 91.5 & 93.7 &    93.2 & 94.9    & 94.6\\
        &  XLM-R\textsubscript{large} & {92.6} &    {92.1} & {94.2}    & {93.8} & {95.5} & {95.1} \\
         \cline{2-8} 
         &  RoBERTa\textsubscript{base} & 88.7 &    88.3 & 91.8    & 91.6 & 93.7 &    93.5\\
        &  XLM-R\textsubscript{base} & \textbf{90.4}    & \textbf{90.3} & 92.8 &    92.6 & 94.7    & 94.3\\
        &  BERTweet & 90.1 &    89.5 & \textbf{94.1}    & \textbf{93.4} & \textbf{95.2}    & \textbf{94.7}\\
        \hdashline
        \multicolumn{2}{l|}{DCNN (\citeauthor{gui-etal-2018-transferring})} & \multicolumn{2}{l|}{89.9}  &  \multicolumn{2}{l|}{\_}  &  \multicolumn{2}{l}{\_}  \\
        \multicolumn{2}{l|}{DCNN  (\citeauthor{gui-etal-2018-transferring})} & \multicolumn{2}{l|}{91.2 [+a]}  &  \multicolumn{2}{l|}{92.4 [+a+b]}  &  \multicolumn{2}{l}{\_}  \\
         \multicolumn{2}{l|}{TPANN} & \multicolumn{2}{l|}{90.9 [+a]}  &  \multicolumn{2}{l|}{92.8 [+a+b]}  &  \multicolumn{2}{l}{\_}  \\
         \multicolumn{2}{l|}{ARKtagger} &  \multicolumn{2}{l|}{90.4}  &  \multicolumn{2}{l|}{93.2 [+b]}  &  \multicolumn{2}{l}{94.6 [+c]} \\
         \multicolumn{2}{l|}{BiLSTM-CNN-CRF} &  \multicolumn{2}{l|}{\_}  &  \multicolumn{2}{l|}{\_}  &  \multicolumn{2}{l}{92.5 [+c]} \\
         \hline   
    \end{tabular}
    }
    \caption{POS tagging accuracy results on the Ritter11-T-POS (Ritter11), ARK-Twitter (ARK) and \textsc{Tweebank-v2} (TB-v2) test sets. Result of ARKtagger \cite{owoputi-etal-2013-improved} on Ritter11 is reported in the TPANN paper \cite{gui-etal-2017-part}. Note that Ritter11 uses  Twitter-specific POS tags for retweeted (RT), user-account, hashtag and url word tokens which can be tagged perfectly using some simple regular expressions. Therefore, we follow  \newcite{gui-etal-2017-part} and  \newcite{gui-etal-2018-transferring} to tag those words appropriately for all models. Results of ARKtagger and BiLSTM-CNN-CRF \cite{ma-hovy-2016-end} on TB-v2 are reported by \newcite{liu-etal-2018-parsing}. Also note that  ``+a'', ``+b'' and  ``+c'' denote  the additional use of extra training data, i.e. models trained on bigger training data. ``+a'': additional use of the POS annotated data from the English WSJ Penn treebank sections 00-24 \citep{Marcus93building}. ``+b'': the use of both training and validation sets for learning models. % (i.e. here, whole file \texttt{oct27.conll}). 
    ``+c'': additional use of the POS annotated data from the UD\_English-EWT training set \citep{silveira14gold}.}
    \label{tab:posresults}
\end{table}

%NOTE that \footnote{In our preliminary experiments, .}  

 \begin{table}[!t]
    \centering
\resizebox{7.5cm}{!}{
\setlength{\tabcolsep}{0.4em}
    \begin{tabular}{ll|l|l|l|l|l|l}
    \hline
       \multicolumn{2}{c|}{\multirow{3}{*}{\bf Model}} & \multicolumn{2}{c|}{\textbf{WNUT16}} & \multicolumn{4}{c}{\textbf{WNUT17}}\\
         \cline{3-8} 
         & & \multirow{2}{*}{soft} & \multirow{2}{*}{hard}& \multicolumn{2}{c|}{entity} & \multicolumn{2}{c}{surface} \\
         \cline{5-8}
         & &  &  & soft & hard & soft & hard\\
         \hline
         
         \multirow{5}{*}{\rotatebox[origin=c]{90}{Our results}} 
         &  RoBERTa\textsubscript{large} & 55.4 &    54.8 & 56.9    & 57.0 & 55.6 &    55.6\\
        &  XLM-R\textsubscript{large} & 55.8 &    55.3 & 57.1 &    57.5 & 55.9    & 56.4\\
         \cline{2-8} 
         & RoBERTa\textsubscript{base} & 49.7    & 49.2 & 52.2 & 52.0 & 51.2 & 51.0 \\
        &  XLM-R\textsubscript{base} & 49.9    & 49.4 & 53.5 &    53.0 & 51.9    & 51.6\\
        &  BERTweet & \textbf{52.1} &    \textbf{51.3} & \textbf{56.5}    & \textbf{55.6} & \textbf{55.1} &    \textbf{54.1}\\
        \hdashline
        \multicolumn{2}{l|}{CambridgeLTL} & \multicolumn{2}{c|}{52.4 [+b]}  &  \multicolumn{2}{c|}{\_}  &  \multicolumn{2}{c}{\_}  \\
        %\multicolumn{2}{l|}{BiLSTM-CNN-CRF} & \multicolumn{2}{c|}{45.0 [+b]}  &  \multicolumn{2}{c|}{35.2}  &  \multicolumn{2}{c}{\_}  \\
        \multicolumn{2}{l|}{DATNet (\citeauthor{zhou-etal-2019-dual})} & \multicolumn{2}{c|}{53.0 [+b]}  &  \multicolumn{2}{c|}{42.3}  &  \multicolumn{2}{c}{\_}  \\
        \multicolumn{2}{l|}{\newcite{aguilar-etal-2017-multi}} & \multicolumn{2}{c|}{\_}  &  \multicolumn{2}{c|}{41.9}  &  \multicolumn{2}{c}{40.2}  \\
        \hline
    \end{tabular}
    }
    \caption{F1 scores on the WNUT16 and WNUT17 test sets. CambridgeLTL result is reported by \newcite{limsopatham-collier-2016-bidirectional}. ``entity'' and ``surface'' denote the scores computed for the standard entity level and the surface level  \citep{derczynski-etal-2017-results}, respectively. %BiLSTM-CNN-CRF results are reported in the DATNet paper \cite{zhou-etal-2019-dual}. 
    }
    \label{tab:nerresults}
\end{table}

 \begin{table}[!t]
    \centering
\resizebox{7.5cm}{!}{
\setlength{\tabcolsep}{0.4em}
    \begin{tabular}{ll|l|l|l|l|l|l}
    \hline
       \multicolumn{2}{c|}{\multirow{2}{*}{\bf Model}} & \multicolumn{2}{c|}{{AvgRec}} & \multicolumn{2}{c|}{{F\textsubscript{1}$^{\text{NP}}$}}& \multicolumn{2}{c}{{Accuracy}} \\
         \cline{3-8} 
         & & soft & hard & soft & hard & soft & hard\\
         \hline
         \multirow{5}{*}{\rotatebox[origin=c]{90}{Our results}} 
        &  RoBERTa\textsubscript{large} & 72.5 &    72.2 & 72.0    & 71.8 & 70.7    & 71.3\\
        &  XLM-R\textsubscript{large} & 71.7 &    71.7 & 71.1    & 70.9 & 70.7 & 70.6 \\
         \cline{2-8} 
        &  RoBERTa\textsubscript{base} & 71.6    & 71.8 & 71.2&71.2 & 71.6 & 70.9 \\
        &  XLM-R\textsubscript{base} & 70.3 &    70.3 & 69.4    & 69.6 & 69.3 &    69.7\\
        &  BERTweet & \textbf{73.2} &    \textbf{72.8} & \textbf{72.8}    & \textbf{72.5} & \textbf{71.7}    & \textbf{72.0}\\
        \hdashline
        \multicolumn{2}{l|}{\newcite{cliche-2017-bb}} & \multicolumn{2}{c|}{68.1}  &  \multicolumn{2}{c|}{68.5}  &  \multicolumn{2}{c}{65.8}  \\
        \multicolumn{2}{l|}{\newcite{baziotis-etal-2017-datastories-semeval}} & \multicolumn{2}{c|}{68.1}  &  \multicolumn{2}{c|}{67.7}  &  \multicolumn{2}{c}{65.1}  \\
        \hline
        
    \end{tabular}
    }
    \caption{Performance scores on the SemEval2017-Task4A test set. See \newcite{rosenthal-etal-2017-semeval} for the definitions of the AvgRec and F\textsubscript{1}$^{\text{NP}}$ metrics, in which AvgRec is the main ranking metric. }%[+] denotes the use of additional training data.} 
    \label{tab:sem17results}
\end{table}

\begin{table}[!t]
    \centering
\resizebox{7.5cm}{!}{
    \begin{tabular}{ll|l|l|l|l}
    \hline
       \multicolumn{2}{c|}{\multirow{2}{*}{\bf Model}} & \multicolumn{2}{c|}{{F\textsubscript{1}$^{\text{pos}}$}} & \multicolumn{2}{c}{Accuracy} \\
         \cline{3-6} 
         & & soft & hard & soft & hard  \\
         \hline
         \multirow{5}{*}{\rotatebox[origin=c]{90}{Our results}} 
         &  RoBERTa\textsubscript{large} & 73.2 &    71.9 & 76.5    & 75.1\\
        &  XLM-R\textsubscript{large} & 70.8 &    69.7 & 74.2    & 73.2\\
         
         \cline{2-6} 
        & RoBERTa\textsubscript{base} &  71.0    & 71.2 & 74.0 &     74.0\\
        &  XLM-R\textsubscript{base} & 66.6    & 66.2 & 70.8 & 70.8\\
        &  BERTweet & \textbf{74.6}    & \textbf{74.3} & \textbf{78.2}    & \textbf{78.2}\\
        \hdashline
        \multicolumn{2}{l|}{\newcite{wu-etal-2018-thu}} & \multicolumn{2}{c|}{70.5}  &  \multicolumn{2}{c}{73.5}  \\
        \multicolumn{2}{l|}{\newcite{baziotis-etal-2018-ntua-slp-semeval}} & \multicolumn{2}{c|}{67.2}  &  \multicolumn{2}{c}{73.2}  \\
        \hline
    \end{tabular}
    }
    \caption{Performance scores on the SemEval2018-Task3A test set. F\textsubscript{1}$^{\text{pos}}$---the main ranking metric---denotes the F\textsubscript{1} score computed for the positive label.}
    \label{tab:sem18results}
\end{table}

 \section{Experimental results}\label{sec:results}
 
\subsubsection*{Main results}

Tables \ref{tab:posresults}, \ref{tab:nerresults}, \ref{tab:sem17results} and \ref{tab:sem18results} present our obtained scores for  BERTweet and baselines regarding  both ``soft'' and ``hard'' normalization strategies. 
We find that for each pre-trained language model the ``soft'' scores are generally higher than the corresponding ``hard'' scores, i.e. applying lexical normalization dictionaries to normalize word tokens in Tweets generally does not help improve the performance of the pre-trained language models on downstream tasks. 

Our BERTweet outperforms its main competitors RoBERTa\textsubscript{base}  and XLM-R\textsubscript{base} on all experimental datasets (with only one exception that XLM-R\textsubscript{base} does slightly better than BERTweet on Ritter11-T-POS). Compared to  RoBERTa\textsubscript{large}  and XLM-R\textsubscript{large} which use significantly larger model configurations, we find that they obtain better POS tagging and NER scores than BERTweet. However, BERTweet performs better than those large models on the two text classification datasets.

Tables \ref{tab:posresults}, \ref{tab:nerresults}, \ref{tab:sem17results} and \ref{tab:sem18results}  also compare our obtained scores with the previous highest reported results on the same test sets. Clearly, the pre-trained language models help achieve new SOTA results on all experimental datasets. Specifically, BERTweet improves the previous SOTA in the novel and emerging entity recognition by absolute 14+\% on the WNUT17 dataset, and in text classification  by 5\% and 4\%  on the SemEval2017-Task4A  and SemEval2018-Task3A  test sets, respectively. 
Our results confirm the effectiveness of the large-scale BERTweet for Tweet NLP.

\subsubsection*{Discussion}

Our results comparing the ``soft'' and ``hard'' normalization strategies with regards to the pre-trained language models confirm the previous view that lexical normalization on Tweets is a lossy translation task \cite{owoputi-etal-2013-improved}. 
We find that RoBERTa outperforms XLM-R on the text classification datasets. This finding is similar to what is found in the XLM-R paper \citep{conneau2019unsupervised} where XLM-R obtains lower performance scores than RoBERTa for sequence classification tasks on traditional written English corpora.
 
We also recall that although RoBERTa and XLM-R use 160 / 80 = 2 times and 301 / 80 $\approx$ 3.75 times bigger English data than our BERTweet, respectively, BERTweet does better than its competitors RoBERTa\textsubscript{base} and XLM-R\textsubscript{base}. Thus this confirms the effectiveness of a large-scale and domain-specific pre-trained language model for English Tweets. In future work, we will release a ``large'' version of BERTweet, which possibly performs better than RoBERTa\textsubscript{large} and XLM-R\textsubscript{large} on all three evaluation tasks. 

%\section{BERTweet-COVID19: Pre-trained language models for analyzing COVID-19 English Tweets}

\section{Conclusion}
We have presented the first large-scale language model BERTweet pre-trained for English Tweets. We demonstrate the usefulness of BERTweet by showing that  BERTweet outperforms its baselines RoBERTa\textsubscript{base} and XLM-R\textsubscript{base} and helps produce better performances than the previous SOTA models for three downstream Tweet NLP tasks of POS tagging,  NER, and text classification (i.e. sentiment analysis \& irony detection). %By publicly releasing BERTweet, %\footnote{\url{https://github.com/VinAIResearch/BERTweet}} 
%we hope that it can foster future research and applications of Tweet analytic tasks.  

As of September 2020, we have collected a  corpus of about 23M ``cased'' COVID-19 English Tweets  consisting of at least 10 and  at  most  64  word tokens. In addition, we also create an ``uncased'' version of this corpus. Then we continue pre-training from our pre-trained BERTweet on each of the ``cased'' and ``uncased'' corpora of 23M Tweets for  40 additional epochs, resulting in two BERTweet variants of pre-trained ``cased'' and ``uncased'' \emph{BERTweet-COVID19} models, respectively. 
By publicly releasing BERTweet and its two variants,  we hope that they can foster future research and applications of Tweet analytic tasks, such as identifying informative COVID-19 Tweets \cite{covid19tweet} or extracting COVID-19 events from Tweets \cite{zong2020extracting}.
 
{%\footnotesize
\bibliographystyle{acl_natbib}
\bibliography{REFs}
}

\end{document}